\documentclass[pmlr]{jmlr}

\usepackage{longtable}
\newcommand{\bftab}{\fontseries{b}\selectfont}
\usepackage[ruled,vlined, noend]{algorithm2e}

\usepackage{booktabs}
\usepackage[load-configurations=version-1]{siunitx} 

\makeatletter
\def\set@curr@file#1{\def\@curr@file{#1}} 
\makeatother

\theorembodyfont{\upshape}
\theoremheaderfont{\scshape}
\theorempostheader{:}
\theoremsep{\newline}

\jmlrvolume{149}
\jmlryear{2021}
\jmlrworkshop{Machine Learning for Healthcare}

\title[CheXbreak]{CheXbreak: Misclassification Identification for Deep Learning Models Interpreting Chest X-rays}

\author{\Name{Emma Chen}
    \thanks{These authors contributed equally to this research.}
      \Email{ychen421@cs.stanford.edu}
      \AND
      \Name{Andy Kim}
      \footnotemark[1]
      \Email{andydhk@cs.stanford.edu}
      \AND
      \Name{Rayan Krishnan}
      \footnotemark[1]
      \Email{rayank@cs.stanford.edu} 
      \AND
      \Name{Jin Long}
      \Email{jinlong@stanford.edu} 
      \AND
      \Name{Andrew Y. Ng}
      \Email{ang@cs.stanford.edu} 
      \AND
      \Name{Pranav Rajpurkar}
      \Email{pranavsr@cs.stanford.edu }
      \\
      {\it Stanford University, USA}
      }

\begin{document}
\maketitle

\begin{abstract}
A major obstacle to the integration of deep learning models for chest x-ray interpretation into clinical settings is the lack of understanding of their failure modes. In this work, we first investigate whether there are patient subgroups that chest x-ray models are likely to misclassify. We find that patient age and the radiographic finding of lung lesion, pneumothorax or support devices are statistically relevant features for predicting misclassification for some chest x-ray models. Second, we develop misclassification predictors on chest x-ray models using their outputs and clinical features. We find that our best performing misclassification identifier achieves an AUROC close to 0.9 for most diseases. Third, employing our misclassification identifiers, we develop a corrective algorithm to selectively flip model predictions that have high likelihood of misclassification at inference time. We observe F1 improvement on the prediction of Consolidation (0.008 [95\% CI 0.005, 0.010]) and Edema (0.003, [95\% CI 0.001, 0.006]). By carrying out our investigation on ten distinct and high-performing chest x-ray models, we are able to derive insights across model architectures and offer a generalizable framework applicable to other medical imaging tasks.
\end{abstract}

\section{Introduction}

There have been significant advancements in developing deep learning algorithms for automatic chest x-ray interpretation (\cite{nam_development_2019, singh_deep_2018,rajpurkar2020chexpedition}) to improve worldwide access to radiology expertise (\cite{moriarity_work_2014, european_society_of_radiology_esr_european_2016, joarde_chest_2009}). The CheXpert competition (\cite{irvin2019chexpert}) has attracted more than 158 research teams over a span of two years from January 2019, with the best performing models achieving an AUROC of 0.930, higher than the reported radiologist benchmark. However, while these models make accurate predictions on most x-rays drawn from the same distribution, a major barrier to the adoption of chest x-ray models to clinical settings is a lack of understanding about their failures. For example, models may be prone to misclassification on patients with certain radiographic findings and clinical features. As another example, low confidence model outputs might be associated with more mistakes, but these associations have not been well explored. Efforts to address these problems by identifying patient subgroups and chest x-ray images with high likelihood of misclassification may help protect against inaccurate diagnoses, determine when clinical decisions should be deferred to downstream experts, and enable monitoring of deep learning based models in clinical workflow (\cite{pmlr-v119-mozannar20b, oakdenrayner2019hidden}).



\begin{figure}
  \includegraphics[width=\linewidth]{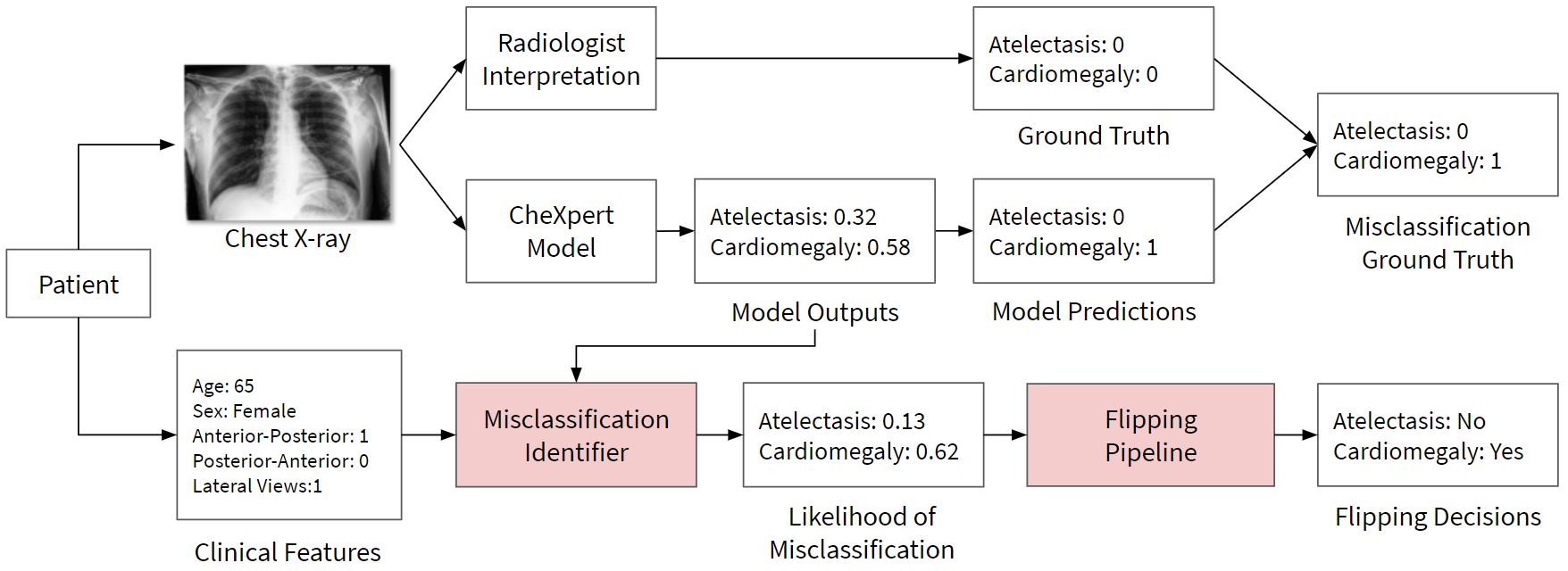}
  \caption{We build misclassification identifiers based on outputs of top CheXpert models and clinical features available for each study. These identifiers are then used to selectively flip results for performance improvement.}
\end{figure}

\subsection*{Generalizable Insights about Machine Learning in the Context of Healthcare}


Our work offers a generalizable framework to many different tasks that may help mitigate and identify failures with model deployment. We first identify patient subgroups with a higher likelihood of misclassification. Knowledge of these subgroups may provide relevant information for deploying these models and assist in clinical decision making. Second, we perform an investigation on ten high-performing chest x-ray models to develop a framework that can identify misclassification with additional data available at inference time. By iterating our methodology with varying models, we are  able to derive insights across different model architecture and training mechanisms. Third, we develop a corrective algorithm that can help improve performance of these chest x-ray models at inference time, i.e. without retraining. This method may demonstrate the possibility for medical machine learning models to use additional data available after training time to improve quality of model prediction. We hope that such efforts will help facilitate trust and improve safety of AI-assistance in the clinical workflow.

\section{Related Work}
Previous work on misclassification identification has been carried out by auditing a model's prediction and determining whether or not to reject it. Some studies have sought to calculate a numerical value that determines the trustworthiness of a model’s output (\cite{schulam2019trust, lakshminarayanan2017simple, pmlr-v48-gal16, kendall2017uncertainties}). For example, Jiang et al. proposed the use of a ``trust score", calculated as the ratio between the distance from a given sample to the nearest but not the predicted class and the distance to the predicted class, to estimate the trustworthiness of a prediction (\cite{jiang2018trust}). Hendrycks \& Gimpel showed that softmax probability could be used as a score to estimate the correctness of a prediction with reasonably good performance (\cite{hendrycks_baseline_2018}). Others have introduced frameworks to train predictive models that are allowed to abstain whenever they are not sufficiently confident in their prediction (\cite{Thulasidasan, pmlr-v97-geifman19a, breiman_randomizing_2000, madras2018predict}). For instance, Cortes et al. presented a framework that trains a classifier with a rejection function, which allows the loss of rejection and the loss of non-rejection for wrong prediction to be incorporated into training (\cite{46544})). Similarly, Geifman \& El-Yaniv developed a model that allows end-to-end optimization of both classification and rejection simultaneously (\cite{NIPS2017_4a8423d5}). These methods typically require models to be retrained in order to estimate the likelihood of misclassification and cannot incorporate new information acquired after training time. However, after chest x-ray models are deployed, there may exist useful information, such as medical history and other sensitive medical data, that cannot be obtained by model developers but easily available to the deploying institution. This motivates the development of an algorithm which can incorporate new features without having to retrain the original model at inference time.

\section{Cohort}

\subsection{CheXpert} 
CheXpert (\cite{irvin2019chexpert}) is a large public dataset for chest x-ray interpretation, including over 224,316 chest x-rays of 65,240 patients. The CheXpert competition includes a validation and  test set for official evaluation. In our experiments, we combine the CheXpert validation and test sets, which consists of 200 and 500 studies respectively, to create a dataset of 700 studies with 902 chest x-ray images, each with 5 radiological observations. We use only the validation and test set since they are higher quality as a result of being manually annotated by board-certified radiologists; the training set instead uses an automated rule-based labeler to extract observations from text radiology reports. We also have access to additional clinical features which, combined with the aforementioned radiological observations, creates a total of 14 labels for each image.

\subsection{Top Models on CheXpert Leaderboard} 
We use ten CheXpert-trained models (or \textit{CheXpert models}) in our experiments, each selected sequentially from the top of the leaderboard. However, if a model has the same name or submitter as a previously-selected model, it is skipped to ensure diversification such that our findings apply to different architectures. All of the selected models are ensembles and many utilize Densely Connected Convolutional Networks (\cite{huang2018densely}) as part of their ensemble. 

\begin{table}[htbp]
  \centering 
  \caption{Example of Misclassification Ground Truth for a single study. Model thresholds are calculated by finding a disease-specific value that maximizes the Youden's Index across a subset of the dataset.}
  \begin{tabular}{cccc}
    \toprule
    \quad & Cardiomegaly & Edema & Consolidation \\
    \toprule
    Model Output      & 0.532 & 0.123 & 0.394  \\
    \hline
    Model Threshold   & 0.7   & 0.5   & 0.2     \\
    \hline
    Model Prediction  & 0     & 0     & 1   \\
    \hline
    Ground Truth      & 0     & 1     & 1   \\
    \hline
     \begin{tabular}{@{}c@{}}Misclassification \\ Ground Truth\end{tabular} & 0     & 1     & 0      \\
    \bottomrule
  \end{tabular}
  \label{tab:example} 
\end{table}





\section{Model Misclassification Based on Clinical Features}

\subsection{Method}
In a real-world clinical setting, radiologists have access to patient data other than x-ray scans, such as patient demographics, medical history and other medical conditions, to help with diagnosis (\cite{boonn_radiologist_2009, leslie_influence_2000}). It has been shown in other statistical models used for medical diagnosis that these clinical features improve diagnostic performance  (\cite{ustun_clinical_2016}). These findings suggest that clinical features may also be significant for training misclassification identifiers.

Clinical features are information about the study (e.g. patient age and sex) which may not always be available at training time; that is, they are typically confidential information that model developers may not have access to in the training phase. This means that hospitals will need to adapt these models to include private health data for the best deployment performance.

Even if these clinical features are widely available, they may not be used by developers in training their models. For instance, developers who made the state-of-the-art CheXpert models did not make use of any provided clinical features. This distinction of parties between the developer and the hospital suggests that there will likely always be different data available during training and deployment. 

We seek to determine whether clinical features are valuable for predicting misclassification. To determine whether features are valuable, we build models using available clinical information and test to see whether any are statistically predictive. The benefits of including clinical features can be extended to other features that are not investigated in this study.

We develop logistic regression models (using the \texttt{statsmodel} library ( \cite{seabold2010statsmodels})) on available \textit{clinical} features -- age, sex, presence of lateral view, number of AP views, number of PA views -- to predict the probability of misclassification and evaluate which features provide new meaningful information to the models. A logistic regression model is constructed for each of the ten models and for each of the five tasks, resulting in a total of 50 statistical models. 

The statistical significance of each feature is evaluated over all the statistical models by considering the number of models for which the feature is significant for a given task. Statistical significance is demonstrated when a clinical feature exhibits a p-value less than 0.05. Next, we compute the average odds ratio across models of the same task and calculate their respective 95\% confidence intervals. The odds ratio is computed by taking the natural exponential of the linear model's coefficients. 

To obtain the misclassification ground truth, we randomly select 504 studies of the combined CheXpert validation and test sets as a ``training" set and find a prediction threshold that maximizes the Youden’s Index on each disease. This threshold is then used to binarize outputs of all models across all 700 studies.

\begin{figure*}[t]
\centering
  \includegraphics[width=0.6\linewidth]{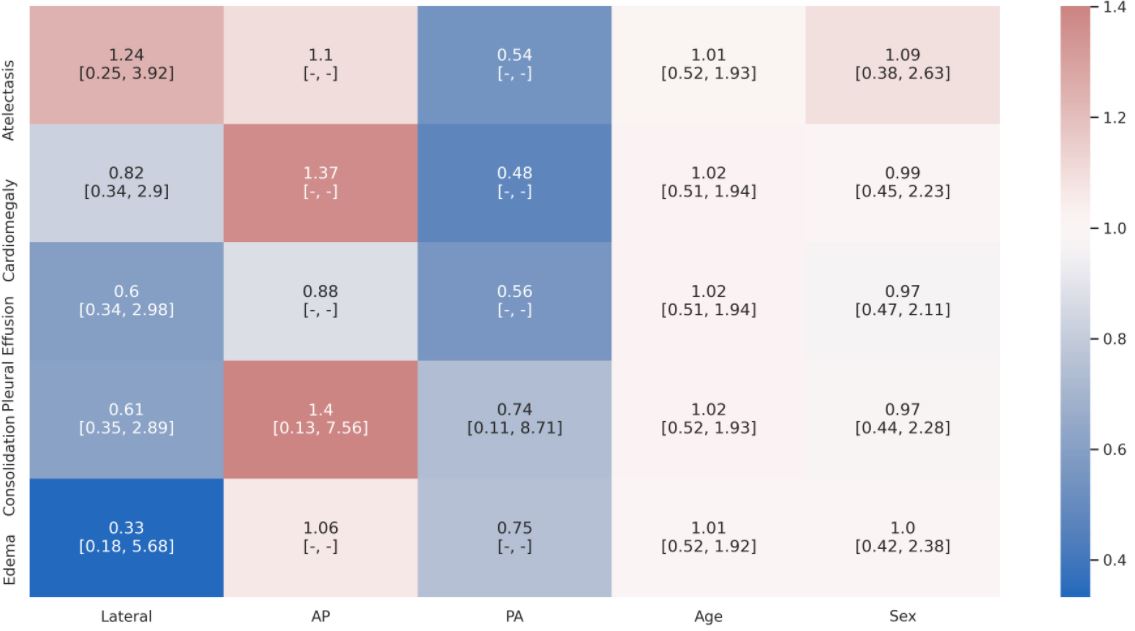}
  \caption{Average odds ratios for models that use a given feature to predict one of the five diseases. ``-" indicates a very large confidence interval that we shorten to prevent overflow.}
\end{figure*}

\subsection{Results}
We find that age is a significant predictor of misclassification for Atelectasis on five models, for Cardiomegaly on seven models, for Pleural Effusion on five models, for Consolidation on five models and for Edema on two models. Age was a significant feature by p-value for all tasks with average odds ratio over 1 (Atelectasis 1.011 [95\% CI 0.517, 1.935], Cardiomegaly 1.018 [95\% CI 0.514, 1.944], Pleural Effusion 1.016 [95\% CI 0.514, 1.945], Consolidation 1.016 [95\% CI 0.518, 1.929], Edema 1.008 [95\% CI 0.521, 1.919]). An odds ratio greater than 1 indicates that as age increases, so does the incidence of misclassification. Because the aggregated confidence intervals for each of these features across all models did not exhibit significance for the odds ratio, we cannot establish a generalized trend for the presence/absence of these features on the rates of misclassification. This does not deny that the features were important predictors for misclassification for specific models on their tasks as demonstrated by their p-values. 

We also find that the presence of lateral views is a significant predictor of misclassification by p-value for two models on Edema (odds ratio of 0.061 for model 7 and 0.057 for model 6). The presence of lateral views is only statistically significant for the task of predicting Edema for which it has an average odds ratio of 0.333 [95\% CI 0.176, 5.676] for all ten models. Although the presence of lateral views is only a significant feature for two of the ten models, an additional lateral view greatly decrease rates of misclassification on those few models for which it is statistically significant.
None of the other features are statistically significant for any of the models or for any of the tasks.

\subsection{Analysis}
In this experiment, we consider age as an input feature and thereby study whether age is an important feature for predicting misclassification. There is a known positive correlation between age and the probability of comorbidities (\cite{fried_association_1999}). Further, patient comorbidities may result in increased rates of misclassification as a consequence of the visual occlusions of some pathologies or radiologist search-satisfying bias (\cite{evans_why_2019}). It is possible that the presence of other diseases influences the misclassification of a given disease. This potential lurking variable is further explored by building a joint model of patient age and number of comorbidities to see whether they are independently relevant for predicting misclassification. We found that the number of comorbidities feature for all models for all tasks had a p-value greater than 0.05. Because the feature exhibited no significance, this hypothesis is rejected.

\subsection{Discussion}
\begin{figure*}[t]
  \includegraphics[width=1\linewidth]{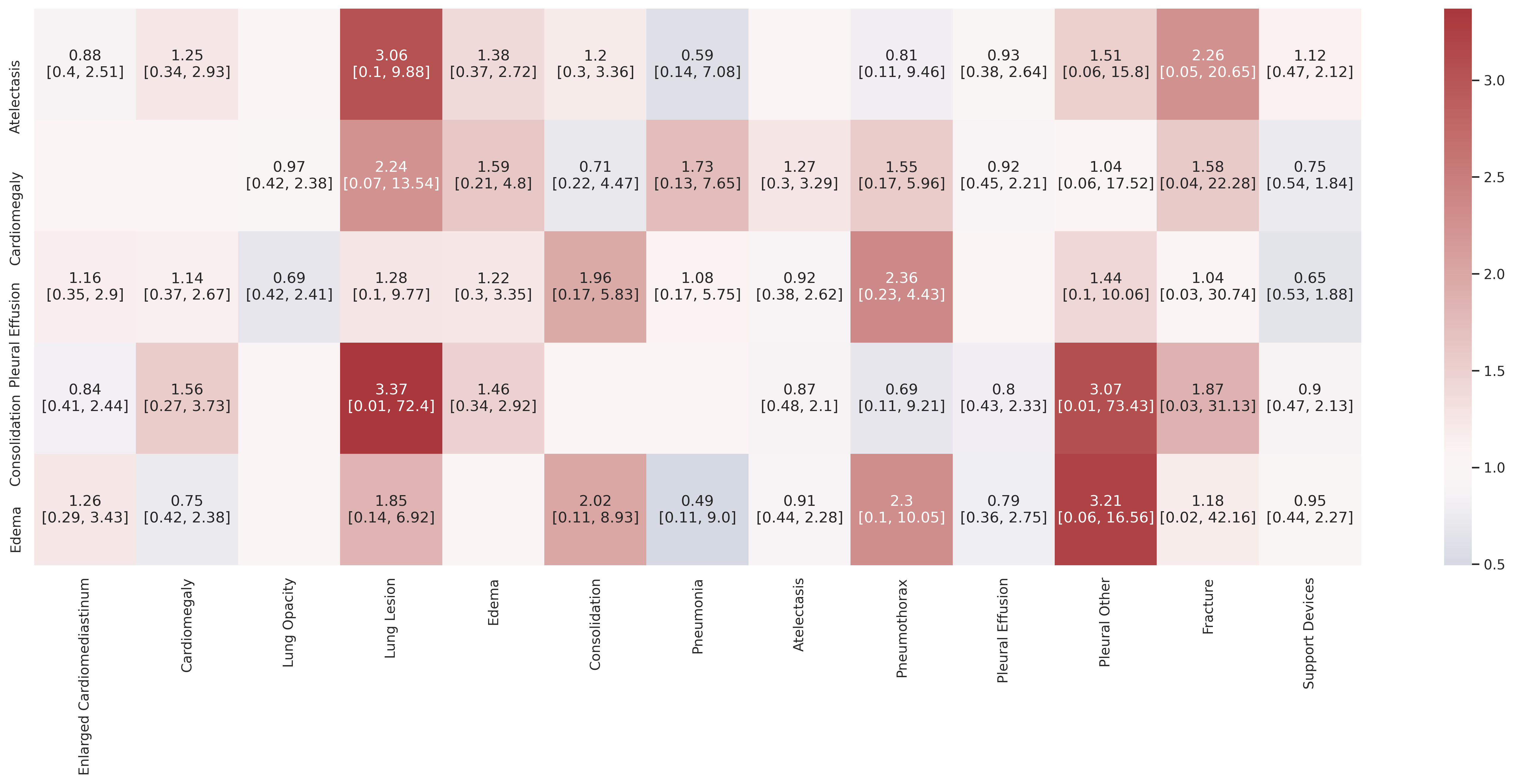}
  \caption{Average odds ratios for models that use a given feature to predict one of the five diseases. The features shown in the column were used to predict the presence/absence of the radiological findings shown on the rows. Empty cells indicate that the given feature was not included in predicting that radiological finding because they are dependent on one another.}
\end{figure*}
The patient's age and presence of lateral views taken of them are statistically significant features by p-value for some models. This result demonstrates that there is value in using clinical features to predict misclassification.

Age was found to be a widely important feature across diseases and across the ten models. We observe that the inclusion of additional lateral views results in lower rates of misclassification of Edema for models 6 and 7. The radiological manifestations of pulmonary edema include peribronchial cuffing and areas of alveolar airspace consolidation (\cite{gluecker_clinical_1999}). It is possible that lateral views present more clear contrasts for identification of Edema rather than that of frontal views, or it may be the case that a model’s joint consideration of an additional lateral view in conjunction with the frontal view results in a better diagnosis. Future work might explore the effective performance of human radiologists in diagnosing patients with Edema to determine if they also improve with an additional lateral view. These results show which models are less prone to misclassification for certain subgroups and may inform how they are interpreted in the clinical setting.

Other studies (\cite{larrazabal_gender_2020,feldman_quantifying_2019}) have reported a significant correlation between the sex balance of the training set with the performance of Convolutional Neural Networks on each sex. They generally observed worse performance for the minority class. This may suggest that knowledge of the patient's sex in an unbalanced dataset might be valuable in predicting misclassification, but we did not find evidence of this in our experiment. The CheXpert dataset has a 59.37\%, 40.63\% split for male and female patients respectively. It is possible that the models investigated here, given their state-of-the-art performance, differed architecturally from those survey models examined in previous work which are more prone to dataset bias. Further, the previous study made use of weak labels (automatic labeler-generated) rather than the verified ground truth here applied in the CheXpert validation and test datasets. Perhaps there is an association between label noise and sex rather than directly between gender and misclassification.


\section{Model Misclassification Based on Relationship to Other Radiological Findings}
\subsection{Method}
We explore whether the occurrence of different diseases can be used to predict misclassification. We hypothesize that the existence of other diseases may help predict the misclassification of a given disease of interest. 

\begin{figure}[h]
  \centering
  \includegraphics[width=0.7\linewidth]{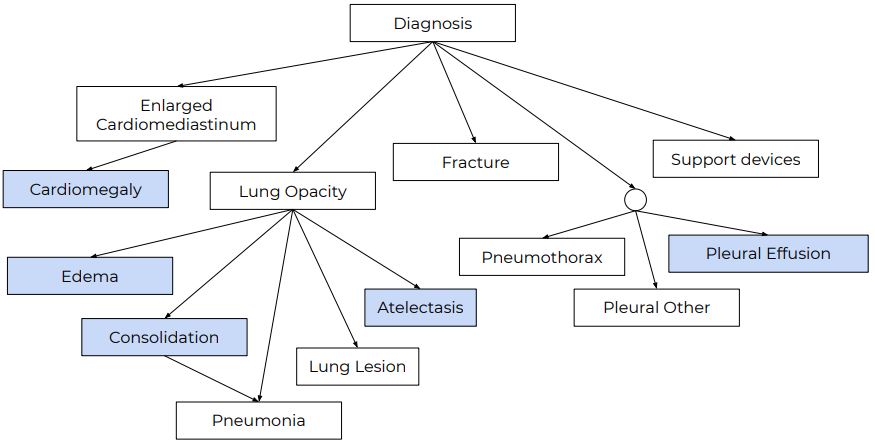}
  \caption{Hierarchy of diseases used in determining independent features. Diseases that models are trained to predict are indicated in blue. Descendants and ancestors of the task disease are excluded as features. As an example, all radiologist diagnoses except for Lung Opacity and Pneumonia are used to predict the misclassification of a model diagnosing Consolidation. }
  \label{fig:3}
\end{figure}

When radiologists labeled the initial studies, they did so by marking the presence or lack of presence of 14 total radiological findings, from which five were selected as CheXpert competition tasks based on their prevalence and clinical importance. Therefore, we have verified radiologist interpretation of the nine other radiological findings unseen by all CheXpert models: No Finding, Enlarged Cardiomediastinum, Lung Opacity, Lung Lesion, Pneumonia, Pneumothorax, Pleural Other, Fracture, and Support Devices. The disease being predicted for and its direct ancestors/descendants, as described in Figure~\ref{fig:3}, are removed from the training set for that disease. The misclassification ground truth is identical as the previous experiment.

A nearly identical procedure from the clinical features experiment is replicated to determine the association between the information about the presence of other diseases with misclassification of a given model on a given task. Here the input features are the ground truth data indicating the presence or absence of all labeled diseases distinct from the one being predicted for. Again, a logistic regression model is trained for each CheXpert model in predicting each disease. The average odds ratios is calculated over all ten models with corresponding confidence intervals.

\subsection{Results}

We find that the presence or absence of Support Devices is a statistically significant predictor of misclassification for three models detecting Cardiomegaly, for four models detecting Pleural Effusion, and for two models detecting Consolidation. 

Edema is an important feature for two models in predicting misclassification of Cardiomegaly. These models have odds ratios of 2.775 and 2.196. This means that although Edema has a limited impact, its presence results in a significant increase in the likelihood of misclassification for those two models where it is a relevant feature. 

Cardiomegaly is an important feature for two models detecting Pleural Effusion, one model for detecting Consolidation and three models for detecting Edema.

Consolidation is an important feature for one model detecting Pleural Effusion, one model detecting Cardiomegaly and for two models detecting Edema. 

Enlarged Cardiomediastinum is statistically significant for one model detecting Edema and one model detecting Pleural Effusion. Lung lesion is statistically significant for three models detecting Consolidation and one model detecting Atelectasis. Pneumothorax is statistically significant for one model detecting Edema, five models detecting Pleural Effusion and one model detecting Cardiomegaly. 
None of the other features are statistical significance for any model for any task.

Our findings suggest that there is a significant relationship between the presence of some radiological findings and misclassification likelihood of other diseases. The experiment also demonstrates the potential to use the predictions of a model on other radiological findings to predict the misclassification on a single disease at a time.

\section{Misclassification Identification With Model Outputs and Clinical Features}
\subsection{Method}

We investigate whether it is possible, at inference time, to identify misclassifications for a given disease using model outputs of all diseases and clinical features. The experiment is motivated by the idea that (a) model outputs may reflect confidence in its prediction (\cite{10.1145/1102351.1102430}) and (b) data obtained after model development could be a useful resource to predict mistakes.  

We train LightGBM classifiers with three different types of input data and evaluate their performance on the task of predicting misclassification. The first is trained on clinical features only (\textit{``clinical only"}); the second on clinical features and the model output value of the disease of interest (\textit{``same label"}); and the third on clinical features and the model output of all diseases (\textit{``all labels"}). For a given study, each CheXpert model outputs five probability scores that correspond to the likelihood of five different diseases: Cardiomegaly, Edema, Pleural Effusion, Consolidation, and Atelectasis. Thus, when predicting misclassification on Cardiomegaly, ``same label" will be trained on the model output of Cardiomegaly and five clinical features (six total features) while ``all labels" will be trained on model output of all five diseases and five clinical features (ten total features). We call these group of classifiers \textit{misclassification identifiers} to emphasize their ability to output likelihood of misclassification and identify potential misdiagnoses. 

We also hypothesize that output values closer to the prediction threshold should show greater uncertainty than those further from it; that is, the negative absolute distance should be proportional to the likelihood of misclassification. Therefore, as a baseline, we create a simple identifier that uses the negative absolute distance from the prediction threshold to the model output (\textit{``naive"}) directly to predict misclassification. Note that the ``naive" identifier is not trained by any machine learning classifier since its value is used directly as the likelihood of misclassification.

We evaluate each identifier by computing the Area Under the Receiver Operating Characteristic curve (AUROC), averaged over the 10 CheXpert models, for each disease reported with a 95\% confidence interval. This process is described in depth in Appendix \hyperref[appen:B]{B}.

\subsection{Results}
\begin{figure}[ht]
  \centering
  \includegraphics[width=0.6\linewidth]{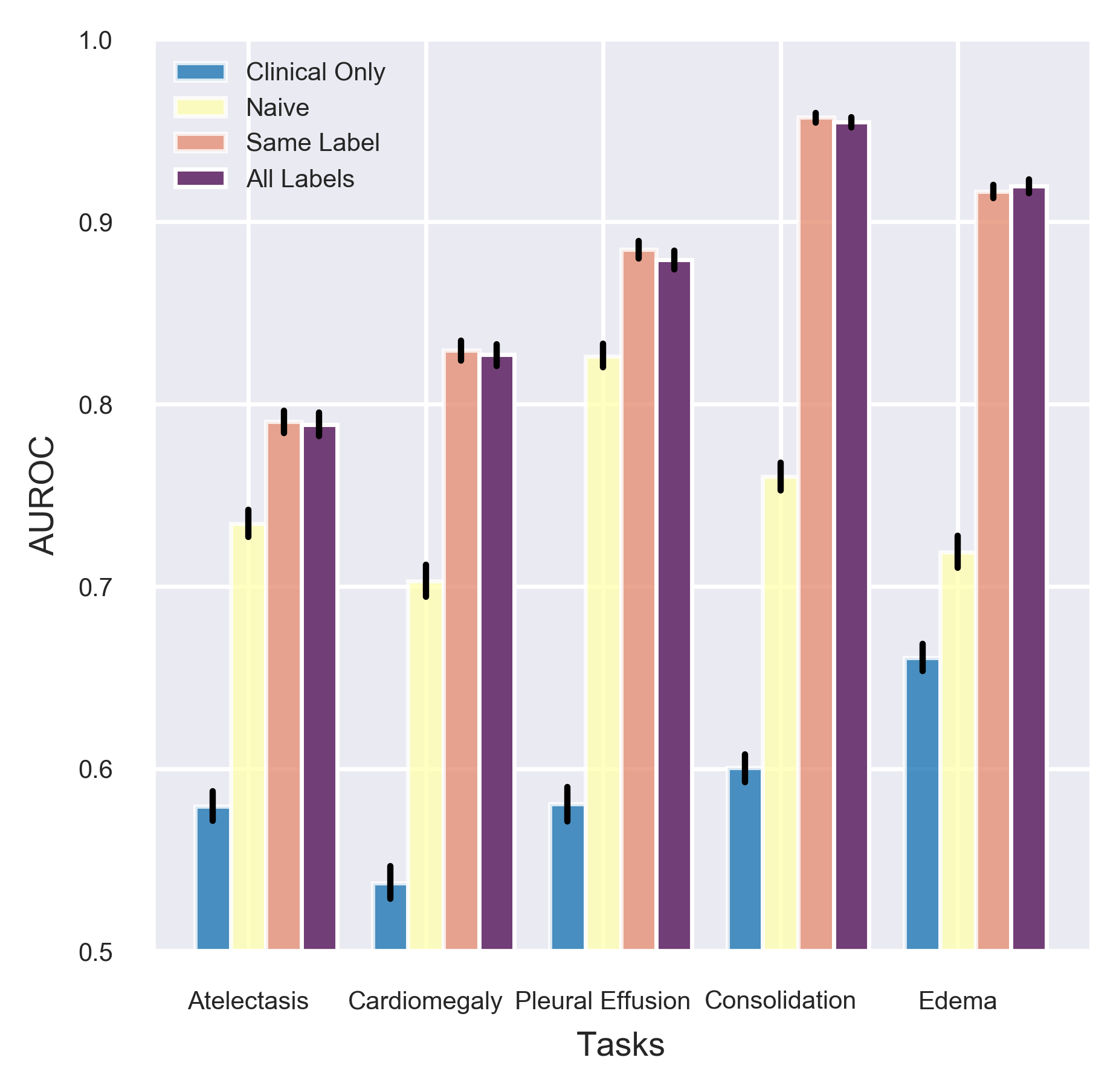}
  \caption{Misclassification Identifier Performance, reported as 95\% confidence intervals averaged over 10 CheXpert models. The ``Same Label" and ``All Labels" models, which use both clinical features and CheXpert model outputs, perform the best with mean AUROCs close to or exceeding 0.9.}
\end{figure}

We find that the ``same label" and ``all labels" identifiers perform very similarly and score the highest AUROC across all diseases, with a mean of 0.881 and 0.880, respectively. On Atelectasis, Cardiomegaly, Pleural Effusion, Consolidation and Edema, we find that the ``same label" mean AUROCs of 0.794, 0.835, 0.892, 0.961, 0.924 respectively, only differ up to 0.007 with the ``all label" mean AUROCs of 0.800, 0.830, 0.885, 0.958, 0.925. Both models achieve a mean AUROC higher than 0.9 on Consolidation and Edema, and a mean AUROC close to 0.9 on Pleural Effusion. Both score a mean AUROC close to 0.8 on Atelectasis and Cardiomegaly. 

Both of these identifiers outperform the ``naive" identifier, with a mean AUROC of 0.751, across all diseases. All of the identifiers outperform the ``clinical only" identifier, with a mean AUROC of 0.595, across all diseases. 

\subsection{Discussion}
In our study, we find that the ``clinical only" identifier, which is independent from model outputs, performs the worst across all diseases. This means that the identifier is not highly discriminative when finding a global relationship between clinical features and misclassification. 

We also find that the ``naive" identifier performs better than the ``clinical only" identifier across all diseases. Its performance as a baseline draws parallels with a former study on multiclass classification problems (\cite{hendrycks_baseline_2018}) which shows that maximum softmax output can be used directly to detect misclassified examples. The main difference is that, in our work, the CheXpert models solve a multi-label binary classification problem and cannot use sigmoid output values directly as a likelihood of misclassification. For instance, in a two-class classification task, the maximum softmax output ranges from 0.5 to 1, with 0.5 denoting high and 1 denoting low likelihood of misclassification. On the other hand, in a perfectly calibrated binary classification task, the sigmoid output ranges from 0 to 1, with both 0 and 1 denoting high confidence for the negative and positive class respectively. Therefore, in the ``naive" identifier, we instead utilize the negative absolute distance from the prediction threshold to the model output. This operation transforms the sigmoid output such that it becomes analogous to the maximum softmax output of a multiclass classification task with two classes, where a higher value indicates a higher likelihood of misclassification. 

The ``same label" and ``all labels" identifiers, which use model outputs along with clinical features, show the best performance among identifiers investigated. The AUROC values are generally close to or exceed 0.9, which demonstrate that high performing misclassification identifiers can be trained by model outputs. We note that both identifiers essentially perform the same; this shows that multi-label does not add value over the single label output, and therefore the inclusion of other model outputs is not additive in predicting misclassification.

Our findings demonstrate that misclassification identifiers can be trained by utilizing clinical features and the sigmoid output value of the disease of interest. In a real-life clinical setting, a high-performance identifier can be used to suggest that a radiologist verify the prediction of a CheXpert model when its likelihood of misclassification is significant. Future work may extend our work to explore these relationships on new test-time distributions (\cite{rajpurkar2020chexphotogenic, phillips2020chexphoto}). To train even better performing identifiers, it may be beneficial to examine other data available during inference time that help predict misclassification, including but not limited to symptoms, prior medical history, and demographics.

\section{Improving Model Performance by Flipping Predicted Misclassifications}
\subsection{Method}
We investigate whether performance of the original CheXpert models can be improved by ``flipping" model predictions with highest likelihood of misclassification. The flipping process is performed by (1) determining a flipping threshold, (2) obtaining all studies with a higher likelihood of misclassification than the threshold, and (3) flipping the predictions of these studies to the opposite outcome. We define the flipping threshold as the $k^{th}$ highest likelihood of misclassification on a training set, where $k$ is selected among a range of values on the validation set to have the best F1 improvement on disease prediction. The likelihood of misclassification is obtained by using the two best-performing misclassification identifiers from the previous section: ``same labels" and ``all labels". 

We note that the flipping algorithm may worsen the performance of the original model if more correct model predictions are reversed than wrong. Although high-performing misclassification identifiers are beneficial for the flipping process, we find that high AUROC on misclassification prediction does not necessarily result in F1 improvement on disease prediction. As a solution, we propose a flipping rule that can help decide whether a model will benefit from the flipping process. We further investigate these ideas below.

\subsubsection{AUROC of Misclassification Prediction does not Translate into F1 Improvement on Disease Prediction }

We find that using high-performing misclassification identifiers (with AUROC higher than 0.9) does not necessarily translate into improved F1 score on disease prediction. This happens because while the AUROC measures performance of misclassification prediction, the F1 score measures performance of disease predictions; thus they are inherently measuring performance of different tasks. When evaluating the AUROC of a misclassification identifier, whether a study has a disease or not does not affect its value. On the other hand, since true positives influence F1 scores differently than true negatives, correctly predicting presence of the disease on studies that indeed have the disease affects the F1 score differently compared to studies that do not. We provide an example in Appendix \hyperref[appen:A]{A} which shows that the performance of disease prediction is worsened after flipping despite using a high-performing misclassification identifier.


\subsubsection{Confusion Matrices on Flipping}
To present examples and prove the flipping rule, we break down the confusion matrix on flipping into four sub-matrices based on the presence of disease and flipped partition (Table~\ref{table:3}). The set of studies is first divided into two partitions: a ``flipped" partition with studies that have a likelihood of misclassification higher than the threshold, and a ``non-flipped" partition for the remaining studies. Studies that are flipped are predicted as misclassified while others are predicted as correctly classified. We further divide each partition based on the presence of the disease in the study. Hence, depending on whether a study is flipped and has the disease, it is compared with the misclassification ground truth to be placed in one of the four confusion matrices. 
\begin{table}[h]
\centering
\caption{A confusion matrix on flipping can be broken down into four sub-matrices based on presence of disease and partition. Studies that are flipped are predicted as misclassified, while those remaining are predicted as correct. For example, $K_{p}01$ denotes studies with the disease that had their predictions incorrectly flipped.}
\textbf{Flipped partition (studies predicted as misclassified)}

\begin{tabular}{llllll}
  & \multicolumn{2}{c}{No Disease}        &                       & \multicolumn{2}{c}{Disease}        \\
  & Pred. Cor.        &  Pred. Mis.              &                       &  Pred. Cor.        & Pred. Mis.         \\ \cline{2-3} \cline{5-6} 
\multicolumn{1}{l|}{Cor.} & \multicolumn{1}{l|}{0} & \multicolumn{1}{l|}{$K_{n}01$} & \multicolumn{1}{l|}{} & \multicolumn{1}{l|}{0} & \multicolumn{1}{l|}{$K_{p}01$} \\ \cline{2-3} \cline{5-6} 
\multicolumn{1}{l|}{Mis.} & \multicolumn{1}{l|}{0} & \multicolumn{1}{l|}{$K_{n}11$} & \multicolumn{1}{l|}{} & \multicolumn{1}{l|}{0} & \multicolumn{1}{l|}{$K_{p}11$} \\ \cline{2-3} \cline{5-6} 
\end{tabular}

\bigskip

\textbf{Non-flipped partition (studies predicted as correct)} 

\begin{tabular}{llllll}
  & Pred. Cor.        &  Pred. Mis.              &                       &  Pred. Cor.        & Pred. Mis.            \\ \cline{2-3} \cline{5-6} 
\multicolumn{1}{l|}{Cor.} & \multicolumn{1}{l|}{$R_{n}00$} & \multicolumn{1}{l|}{0} & \multicolumn{1}{l|}{} & \multicolumn{1}{l|}{$R_{p}00$} & \multicolumn{1}{l|}{0} \\ \cline{2-3} \cline{5-6} 
\multicolumn{1}{l|}{Mis.} & \multicolumn{1}{l|}{$R_{n}10$} & \multicolumn{1}{l|}{0} & \multicolumn{1}{l|}{} & \multicolumn{1}{l|}{$R_{p}10$} & \multicolumn{1}{l|}{0} \\ \cline{2-3} \cline{5-6} 
\end{tabular}
\label{table:3}
\end{table}

\subsubsection{Flipping rule}
We propose a flipping rule that takes into consideration the presence of disease in addition to the performance of the misclassification identifier. Since we want to (a) flip more wrong model predictions than correct and (b) increase true positive of disease predictions as much as possible, the flipping rule considers both the top-k precision of the misclassification identifier, as well as the performance of misclassification for studies with disease. The flipping rule states that \textit{we flip data if both $K_{n}11 + K_{p}11 > K_{n}01 + K_{p}01$ (i.e. top-k precision $>$ 0.5) and $K_{p}11 \geq K_{p}01$}.

We prove that by following the flipping rule, we will get non-negative F1 change if flipping is performed on the same set of data the rule is calculated on. It can also be extended to determine whether another dataset with similar distribution would benefit from the flipping algorithm. We denote the precision, recall and F1 after flipping with $precision^\prime$, $recall^\prime$, and $F1^\prime$. We know the following is true by the definition of precision and recall:
\begin{align*}
    &precision^\prime = \frac{K_{p}11 + R_{p}00}{(K_{p}11 + R_{p}00) + (K_{n}01 + R_{n}10)} \\
    &\frac{1}{precision^\prime} = 1 + \frac{K_{n}01 + R_{n}10}{(K_{p}11 + R_{p}00)}\\
    &recall^\prime = \frac{K_{p}11 + R_{p}00}{(K_{p}11 + R_{p}00) + (K_{p}01 + R_{p}10)} \\
    &\frac{1}{recall^\prime} = 1 + \frac{K_{p}01 + R_{p}10}{K_{p}11 + R_{p}00} \\
    &\frac{2}{F1^\prime} = \frac{1}{precision^\prime} + \frac{1}{recall^\prime} = 2 + \frac{K_{n}01 + K_{p}01 + R_{n}10 + R_{p}10}{K_{p}11 + R_{p}00}
\end{align*}

Similarly,
\[\frac{2}{F1} = 2 + \frac{K_{n}11 + K_{p}11 + R_{n}10 + R_{p}10}{K_{p}01 + R_{p}00}\]

Per our assumption, we have that $K_{n}11 + K_{p}11 > K_{n}01 + K_{p}01$ and $K_{p}11 \geq K_{p}01$. Therefore, $\frac{2}{F1^\prime} < \frac{2}{F1}$, which proves that $F1^\prime > F1$. The flipping algorithm is described in depth in Appendix \hyperref[appen:C]{C}.

\begin{figure}[t]
  \centering
  \includegraphics[width=\linewidth]{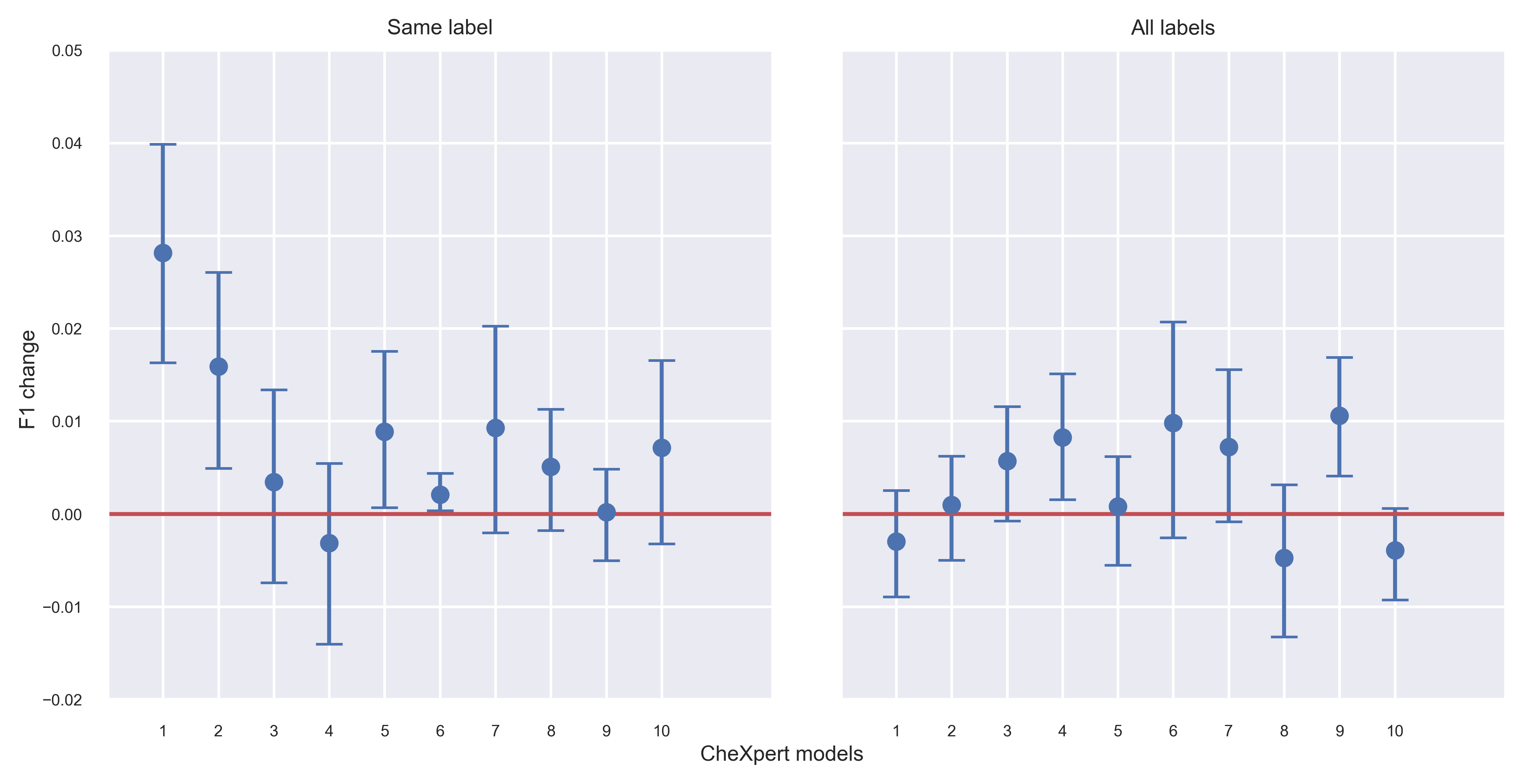}
  \caption{F1 change for CheXpert models after flipping on Consolidation prediction  with the ``same label" (left) and on Edema with ``all labels" (right) identifier. For ``same label", the mean F1 change for eight models are above 0, with three models showing statistically significant F1 improvement and none showing statistically significant F1 decrease.
  For ``all labels", the mean F1 change for seven models are above 0, with two models showing statistically significant F1 improvement and none showing statistically significant F1 decrease.
  }
  \label{fig:5}
\end{figure}

\subsection{Results}

When flipping with the ``same label" misclassification identifier, we see a statistically significant F1 improvement on Consolidation (0.008, [95\% CI 0.005, 0.010]) as well as on the mean of all diseases (0.002, [95\% CI 0.001, 0.002]) averaged on all CheXpert models; when flipping with the ``all labels" misclassification identifier, we see statistically significant F1 improvement on Edema (0.003, [95\% CI 0.001, 0.006]) averaged on all models (Table~\ref{table:5}). All confidence intervals have non-negative upper bound, which indicates that there is no statistically significant decrease in F1 score after flipping. 

Figure~\ref{fig:5} shows the F1 change for each of the ten CheXpert models (with 95\% confidence interval) on diseases that have statistically significant averaged F1 improvement across all models, i.e.  Consolidation for the ``same label" misclassification identifier (left) and Edema for the ``all labels" misclassification classifier (right). For ``same label", we find eight models with mean F1 improvement, three of them being statistically significant. For ``all labels", we find seven models with mean F1 improvement, two of them being statistically significant. We note that while the two misclassification identifiers have almost the same performance in terms of AUROC, ``same labels" has better flipping performance compared to ``all labels".

\subsection{Discussion}
In this experiment, we investigate the use of misclassification identifiers to improve model performance. We show that flipping with the ``same label" identifier on Consolidation, as well as flipping with ``all labels" on Edema, has statistically significant improvement in F1 improvement.

Moreover, we find that flipping does not cause statistically significant decrease in F1 score, meaning that flipping can improve the performance of models on specific diseases detection without having to significantly sacrifice the performance on others. In addition, for the ``same label" model, we see statistical significant F1 improvement on the mean of all diseases; this signifies that it may be possible to improve model performance as a whole. 


We also observe bootstrapped samples with negative F1 change on disease prediction, despite having applied the flipping rule. This happens since we assume similar data distribution across the train-val-test splits and use the flipping rule on training data as a proxy for test data, while the flipping rule guarantees non-negative F1 change only on the set of data which the rule is calculated on. That is, in our experiments, we use the flipping rule on training data to predict whether the test set would benefit from flipping, find the best threshold on the validation set, and calculate F1 change on the test set. This does not necessarily prevent negative F1 change on the test data; however, given that the lower bounds are generally close to 0, the rule may still help with flipping at a lower risk of getting substantial F1 decrease. 

\begin{table}[t]
\centering
\caption{F1 change averaged over all models (with 95\% confidence interval) after flipping. The ``same label'' identifier shows a statistically significant F1 improvement for Consolidation and for the mean over all diseases. The ``all labels'' identifier shows statistically significant F1 improvement for Edema. The rest of the results have non-negative upper bound which shows that the F1 score does not have a significant decrease on any of the diseases.}
\label{tab:commands}

\sisetup{
  table-align-uncertainty=true,
  separate-uncertainty=true,
}
\renewrobustcmd{\bfseries}{\fontseries{b}\selectfont}
\renewrobustcmd{\boldmath}{}

\begin{tabular}{lcc}
\hline
Disease          & Same Label (95\% CI)          & All Labels (95\% CI)          \\ \hline
Atelectasis      & 0.000 [-0.001, 0.000]         & -0.001 [-0.002,  0.000]        \\
Cardiomegaly     & 0.000 [0.000,\: 0.001]          & \: 0.001 [-0.001, 0.002]         \\
Pleural Effusion & 0.000 [-0.001, 0.000]         & -0.001 [-0.002, 0.001]        \\
Consolidation    & \bftab 0.008 [0.005, \:0.010]  & \: 0.003 [-0.001, 0.006]         \\
Edema            & 0.000 [-0.001, 0.002]         & \: \bftab 0.003 [0.001, \:0.006] \\
Mean             & \bftab 0.002 [0.001, \:0.002] & \: 0.001 [0.000,  0.002]          \\ \hline
\end{tabular}
\label{table:5}
\end{table}

Overall, our results suggest that we can improve model outputs by building misclassification predictors based on the logits and clinical feature and following a corrective algorithm. This shows that, for best performance, models need to be adapted to test-time distribution with the flipping rule before deployment.

\section{Limitations \& Future Work}
Although we show clinical and radiological features that are significant predictors of misclassification by p-values, we are unable to establish a significant odds ratio interval across the ten models. This means that we cannot determine whether the presence or absence (or the increase or decrease) of these features results in greater or lesser rates of misclassification. Further work should aim to remedy this by studying models individually and taking more samples into consideration. 

We combine the CheXpert validation and test sets because both include clinical features and are manually annotated by board-certified radiologists. However, the CheXpert models we evaluate were exposed to the validation set during training, so there may exist discrepancy in performance between these two datasets. This work is also limited by the access to the clinical features collected in this dataset. Future work should explore other clinical features, such as ethnicity or weight, and prior medical history.

Although this work investigates cases in which deep learning models make mistakes, it does not further examine why these cases lead models to fail. Investigating why deep learning models fail requires having access to the models themselves; in our case, we only have access to their classification output. Thus, speculating about the characteristics of x-rays or model architectures that cause misclassification is out of scope of this research which is focused on adapting models for new distributions. An interesting avenue of future work would be providing insight as to why the chest x-ray classifier fails. 

Finally, we do not evaluate our misclassification identifiers on other datasets to demonstrate its generalizability on different data sources. Since this methodology may be applied to other areas of model-based diagnosis, an area of future work can be to evaluate our identifiers and methodology on other medical domains. 

\section{Conclusion}
In this study, we investigated whether we awere able to identify and predict misclassifications for top-performing chest x-ray models. We first identified patient subgroups that were more likely to be misclassified by deep-learning based chest x-ray models. Second, we built a misclassification identifier on the top ten CheXpert models using model outputs and additional data available at inference time. Third, we developed a corrective algorithm that flipped prediction labels of studies with the highest likelihood of misclassification to improve the F1 score of CheXpert models. 

We found characteristics of patient subgroups that are significant predictors of misclassification. Older patients and the presence of lateral views were significant clinical features. We also identified some radiologist findings that influence the likelihood of misclassifiation by these models. The presence of Support Devices was a widely significant feature for predicting misclassification across tasks. Meanwhile, the knowledge of other diseases, namely pneumothorax and lung lesion were important radiological findings. These findings may inform clinicians about the risks of deploying top chest x-ray models in practice. 

We also investigated whether we could predict the likelihood of misclassification of a chest x-ray model at inference time, i.e. without retraining. We incorporated clinical features and all model outputs to produce a misclassification likelihood. We found that we could produce a misclassification identifier with high discrimination on Pleural Effusion, Consolidation and Edema with mean AUROCs of 0.892, 0.961, 0.924 respectively.

We further developed a corrective algorithm and proposed a rule that could suggest whether model performance would be improved by flipping predictions with high likelihood of misclassification. We demonstrated statistically significant F1 improvement on Consolidation (0.008, [95\% CI 0.005, 0.010]) and Edema (0.003, [95\% CI 0.001, 0.006]). This result demonstrated the potential of using additional data available at inference time to improve model prediction.

Our work offers a methodology generalizable to many tasks in medical imaging where additional data unavailable for training may be used to improve model deployment. We investigated not only one model, but ten models developed by different teams, and therefore derived insights about the consistency of our findings across chest x-ray model architectures and training procedures. This framework can also be applied to different tasks in medical imaging at inference time. We hope these efforts will help build trust and enable improved safety of AI-assistance in the medical setting.

\acks{We would like to thank Dr. Anuj Pareek for their review of the manuscript. We would like to thank the Stanford Machine Learning Group (stanfordmlgroup.github.io) for support.}

\bibliography{sample}

\appendix
\newpage
\section*{Appendix A. Example of High AUROC on Misclassification Prediction but Worse F1 on Disease Prediction}
\label{appen:A}
Consider a dataset with 50 studies, where only the predictions of misclassification with probability scores near the flipping threshold are wrong, and those away from the threshold (i.e. having either high or low probability scores) are correct. The breakdown of misclassification predictions is shown in Table~\ref{table:4}. The AUROC of such misclassification predictions is 0.984. Now, assume we flip ten studies with the highest likelihood of misclassification. Among the 50 studies, five actually have the disease ($K_p01 + K_p11 + R_p00 + R_p10$). Among the five studies with the disease, the CheXpert model originally predicts that three have the disease ($K_p01 + R_p00$) and that the other two do not ($K_p11 + R_p10$); however, after flipping, the CheXpert model predicts that only one has the disease ($K_p11 + R_p00$) and that the other four do not ($K_p01 + R_p10$). Among the 45 studies without the disease, the CheXpert model originally predicts that 35 do not have the disease ($K_n01 + R_n00$) and that the other ten do ($K_n11 + R_n10$); after flipping, the CheXpert model predicts that 43 do not have the disease ($K_n11 + R_n00$) and that the other two do ($K_n01 + R_n10$). Although we see performance improvement for studies that do not have the disease, the performance decrease for studies that do have the disease leads to a negative F1 change of -0.083.

\begin{table}[h]
\centering
\caption{An example where using a misclassification identifier with AUROC of 0.984 still results in negative F1 change of -0.083 on disease prediction.}
    \textbf{Flipping partition (studies predicted as misclassified)}
\begin{tabular}{llllll}
                              & \multicolumn{2}{c}{No Disease}        &                       & \multicolumn{2}{c}{Disease}        \\
                              & Pred. Cor.           & Pred. Mis.           &                       & Pred. Cor           & Pred. Mis           \\ \cline{2-3} \cline{5-6} 
\multicolumn{1}{l|}{Cor.} & \multicolumn{1}{l|}{$0$} & \multicolumn{1}{l|}{$K_n01 = 0$} & \multicolumn{1}{l|}{} & \multicolumn{1}{l|}{$0$} & \multicolumn{1}{l|}{$K_p01=2$} \\ \cline{2-3} \cline{5-6} 
\multicolumn{1}{l|}{Mis.} & \multicolumn{1}{l|}{$0$} & \multicolumn{1}{l|}{$K_n11 = 8$} & \multicolumn{1}{l|}{} & \multicolumn{1}{l|}{$0$} & \multicolumn{1}{l|}{$K_p11 = 0$} \\ \cline{2-3} \cline{5-6} 
\end{tabular}

\bigskip

    \textbf{Non-flipping partition (studies predicted as correct)} 

\begin{tabular}{llllll}
                              & Pred. Cor           & Pred. Mis.           &                       & Pred. Cor.           & Pred. Mis.           \\ \cline{2-3} \cline{5-6} 
\multicolumn{1}{l|}{Cor.} & \multicolumn{1}{l|}{$R_n00 = 35$} & \multicolumn{1}{l|}{$0$} & \multicolumn{1}{l|}{} & \multicolumn{1}{l|}{$R_p00 = 1$} & \multicolumn{1}{l|}{$0$} \\ \cline{2-3} \cline{5-6} 
\multicolumn{1}{l|}{Mis.} & \multicolumn{1}{l|}{$R_n10 = 2$} & \multicolumn{1}{l|}{$0$} & \multicolumn{1}{l|}{} & \multicolumn{1}{l|}{$R_p10 = 2$} & \multicolumn{1}{l|}{$0$} \\ \cline{2-3} \cline{5-6} 
\end{tabular}
\label{table:4}

\end{table}

\newpage
\section*{Appendix B. Misclassification Identifier Training and Evaluation}
\label{appen:B}
The following algorithm demonstrates the process of training and evaluating the misclassification identifier that is described in section 6.

\begin{algorithm}[h]
\DontPrintSemicolon
\SetAlgoLined
\KwResult{AUROC of misclassification identifier averaged over CheXpert models with 95\% confidence interval}
 \For{each disease}{
 \For{each CheXpert model} {
 \For{each train-test split} {
 On train fold, find threshold that maximizes Youden's Index\;
 On train and test fold, binarize with threshold and calculate misclassification ground truth\;
 On train fold, train misclassification identifier\;
 On test fold, compute likelihood of misclassification with identifier and compute AUROCs with 1000 bootstrapped samples\;
 }
 }
 }
 \caption{Misclassification Identifier Training and Evaluation}
\end{algorithm}

\newpage
\section*{Appendix C. Flipping Process and Evaluation}
\label{appen:C}
The following algorithm demonstrates the process of flipping and evaluating the flipping algorithm that is described in section 7.

\begin{algorithm*}[h]
\DontPrintSemicolon
\SetAlgoLined
\KwResult{F1 improvement after flipping averaged over CheXpert models with 95\% confidence interval}
 \For{each disease}{
 \For{each CheXpert model} {
 \For{each train-val-test split} {
 On train fold, find threshold that maximizes Youden's Index\;
 On train, val and test fold, binarize with threshold and calculate misclassification ground truth\;
 On train fold, train misclassification identifier\;
 On train, val and test fold, compute likelihood of misclassification with identifier \;
 $flip,\ bestImprovement,\ bestThreshold, \ bestFlip \longleftarrow False,\ 0,\ -1, \ False$\;
 \For{various k} {
     \If{(top-k\ precision $>$ 0.5\ and\ Kp11 - Kp01 $>=$ 0) is True on training fold} {
        $flip \longleftarrow True$ \;
        }
    \Else {
    $flip \longleftarrow False$ \;
    }
    
    On train fold, define $flippingThreshold$ as top-kth element's likelihood of misclassification\;
    \If{$flip$ is True}{
     On val fold, flip predictions that have likelihood of misclassification higher than $flippingThreshold$\;
    On val fold, compute F1 improvement\;
     } 
     \Else
     { 
        F1 improvement is 0 on val fold;
     }
    
    \If{F1 improvement $>$ bestImprovement}{
    $bestImprovement,\ bestThreshold, \ bestFlip  \longleftarrow F1\ improvement,\ flippingThreshold, \ Flip$\; 
    }

    }

 \If{$bestFlip$ is True}{
 On test fold, flip predictions that have likelihood of misclassification higher than $bestThreshold$ and compute F1 improvements with 1000 bootstrapped samples\;
 } 
}}}
 \caption{Flipping Process and Evaluation}
\end{algorithm*}

\end{document}